\documentclass{article}

\usepackage{PRIMEarxiv}

\usepackage[utf8]{inputenc} 
\usepackage[T1]{fontenc}    
\usepackage{hyperref}       
\usepackage{url}            
\usepackage{booktabs}       
\usepackage{amsfonts}       
\usepackage{nicefrac}       
\usepackage{microtype}      
\usepackage{fancyhdr}       
\usepackage{graphicx}       
\graphicspath{{media/}}     

\usepackage{amssymb}
\usepackage{amsmath}
\usepackage{booktabs}
\usepackage{tabularx}
\newcolumntype{Y}{>{\centering\arraybackslash}X}
  
\title{Synthetic Image Learning: Preserving Performance and Preventing Membership Inference Attacks
}

\author{
  Eugenio Lomurno, Matteo Matteucci \\
  Politecnico di Milano \\
  Department of Electronics, Information and Bioengineering\\
  Via Ponzio 34/5, 20133 Milan, Italy\\
  \texttt{\{eugenio.lomurno, matteo.matteucci\}@polimi.it} \\
}

\begin{document}
\maketitle

\begin{abstract}
\noindent Generative artificial intelligence has transformed the generation of synthetic data, providing innovative solutions to challenges like data scarcity and privacy, which are particularly critical in fields such as medicine. However, the effective use of this synthetic data to train high-performance models remains a significant challenge. This paper addresses this issue by introducing Knowledge Recycling (KR), a pipeline designed to optimise the generation and use of synthetic data for training downstream classifiers.
At the heart of this pipeline is Generative Knowledge Distillation (GKD), the proposed technique that significantly improves the quality and usefulness of the information provided to classifiers through a synthetic dataset regeneration and soft labelling mechanism. The KR pipeline has been tested on a variety of datasets, with a focus on six highly heterogeneous medical image datasets, ranging from retinal images to organ scans. 
The results show a significant reduction in the performance gap between models trained on real and synthetic data, with models based on synthetic data outperforming those trained on real data in some cases. Furthermore, the resulting models show almost complete immunity to Membership Inference Attacks, manifesting privacy properties missing in models trained with conventional techniques.
\end{abstract}


\section{Introduction}\label{sec:Introduction}
\noindent The advent of generative deep learning has marked a fundamental technological breakthrough that is rapidly permeating every aspect of society and profoundly affecting the daily lives of every individual. Thanks to this technology, it is now extremely easy to create and interact with high-quality synthetic data, be it images, text, audio or video. This ease of access to artificially generated content makes it increasingly difficult to distinguish between human and algorithmic production.
Meanwhile, the applications and innovations of generative models are expanding at a rapid pace, revolutionising many sectors. The implications of this development are profound: on the one hand, new opportunities are opening up, and on the other, ethical and social challenges are emerging in relation to the use and misuse of such technologies.

\noindent Today, this technological progress raises problems related to the circulation of images or text documents generated by algorithms and presented as the fruit of human labour. However, it also opens the door to a dual use with immense virtuous potential. It is precisely the difficulty of distinguishing between human and algorithmic production that has led to the use of generative models to enrich real data sets and, more recently, to attempts at total replacement to obtain entire synthetic datasets. 

\noindent However, the creation of entirely synthetic datasets is a complex task that requires models capable of generating large amounts of data in a reasonable amount of time, while carefully balancing the quality and variety of the data generated. Indeed, it is known that training models based solely on synthetic data tends to degrade performance compared to those trained on real data~\cite{lampis2023bridging}.
In addition to these aspects, it is crucial to consider an area of growing importance, that of data privacy, both before and after the data has been learned by the models. This is particularly critical expecially in the context of medical data, where the protection of privacy is essential to preserve the relationship of trust between experts and patients. In this scenario, generative technology offers unexploited potential for the secure use of medical data, opening up new opportunities for healthcare research and innovation.

\noindent This paper presents the Knowledge Recycling (KR) strategy, a pipeline that aims to improve the generation of synthetic datasets and the training of downstream classifiers using only synthetic images. First, the generator and an auxiliary classifier, named Teacher Classifier, are trained. The optimal checkpoint of the Generator is determined by training a Student Classifier for each checkpoint. These trainings use the proposed technique of Generative Knowledge Distillation, where the Teacher Classifier generates soft labels for the synthetic images, allowing the Student Classifier to learn about uncertainties and class correlations, thus improving its accuracy in predicting both synthetic and real images.
After identifying the best checkpoint, the generation parameters are optimised by adjusting the size of the synthetic dataset, the frequency of dataset regeneration during Student Classifier training, and the standard deviation of the Generator. Once the optimal Student Classifier training is completed, its resistance to a Membership Inference Attack is evaluated and compared with the one achieved by the Teacher Classifier~\cite{shokri2017membership}.

\noindent This work aims to demonstrate the possibility of obtaining classifiers trained on synthetic data with comparable performance to those trained on real data, while providing superior resistance to Membership Inference Attacks. The main contributions of this research are:
\begin{itemize}
    \item The introduction of Knowledge Recycling, a novel pipeline for optimised generation and use of synthetic data in the context of classifier training.
    \item The development of Generative Knowledge Distillation, a technique that improves the quality of information transferred from synthetic data to classifiers, thereby reducing the performance gap with models trained on real data.
    \item Demonstrate the effectiveness of the proposed pipeline in producing models that are nearly immune to Membership Inference Attacks, resulting in a positive trade-off between performance and resistance to this category of privacy attacks.
\end{itemize}

\section{Related Works}\label{sec:Related}
\noindent The state of the art in image generation is currently contested between the Generative Adversarial Networks (GAN) family and the Denoising Diffusion Probabilistic Models (DDPM)~\cite{goodfellow2014generative,ho2020denoising}.
Although using different mechanisms, both families of models are in fact capable of producing and manipulating images with very high resolution by being conditioned in a variety of ways~\cite{kang2023scaling,podell2024sdxl}.
In parallel to this line of research, which aims to produce high quality single images, a second line of research has been developed with the aim of exploiting this generative power to create fully synthetic datasets or to enrich existing ones. 

\noindent The first attempts in this direction concerned contexts where it is very complex and time-consuming to collect and label new data, such as in the medical field. 
Frid-Adar et al. used GANs to generate synthetic images of liver lesions. This work showed that adding synthetic images to the source dataset improved the performance of classification models for diagnosing liver lesions~\cite{frid2018gan}.
Subsequently, Sedigh et al. and Islam et al. also used GAN models to generate synthetic images of skin cancer and brain PET, respectively. In both cases, enriching the dataset with real images led to improved classification performance~\cite{sedigh2019generating,islam2020gan}.

\noindent In addition, studies have been undertaken not to enrich existing image datasets, but to generate entiraly new ones and evaluate their properties via downstream machine learning problems~\cite{lomurno2022sgde,cazenavette2022dataset,sariyildiz2023fake}. 
From early work, it has become clear that the semantic information contained in synthetic data is not in itself sufficient for a model trained on such data to perform well when making inferences on real data~\cite{ravuri2019seeing}.
Techniques have been developed to make the most of the information that can be extracted from generative models, as well as the potentially unlimited number of images that can be generated.
It has been shown that both recycling the synthetic dataset in the training phase and creating synthetic datasets with higher cardinality than the dataset used to train the generator are very beneficial to performance~\cite{besnier2020dataset,lomurno2024stable}.
Filtering techniques were also proposed to discard synthetic images that were classified incorrectly or with low confidence by an auxiliary classifier~\cite{dat2019classifier}. This also allowed sampling from sparser distributions, further enriching the information of the synthetic datasets~\cite{lampis2023bridging}.

\subsection{Privacy Threats and Countermeasures}\label{subsec:Privacy}
\noindent As research on new models and techniques continues and their applications increase, the attack surface on such models and the importance of privacy protection continue to grow.
Among the most common and popular attacks are the families of Membership Inference Attacks (MIAs), Model Inversion, Model Extraction, and Data Poisoning, which can be applied depending on the context and the ability to access and interact with the attacked model~\cite{shokri2017membership,fredrikson2015model,tramer2016stealing,biggio2012poisoning}.
MIAs are the easiest family of attacks to use, as they can be executed in black-box contexts and from logits alone. Their purpose is to guess whether the sample given as input to the model to be attacked was present in its training set or not. From this attack, and thus once the presence of a particular sample learned by the attacked model is known, it is possible, for example, to refine Model Inversion or Model Extraction attacks, or to proceed with inference attacks aimed at extracting more refined information.

\noindent Many defensive mechanisms have been developed to deal with this threat, many of which rely on relaxed forms of Differential Privacy~\cite{abadi2016deep}. Although such mechanisms are very effective in preventing MIAs, they often require very long training times and lead to performance degradation of the protected model~\cite{lomurno2022utility}. Recently, however, in addition to empirical metrics to measure the trade-off between performance and resistance to MIAs, alternative techniques have been proposed to protect models by adversarial training or individual private training steps instead of the entire training~\cite{lomurno2023discriminative,steinke2024privacy}.

\section{Method}\label{sec:Method}

\begin{figure*}[t]
    \centering
    \includegraphics[width=1\linewidth]{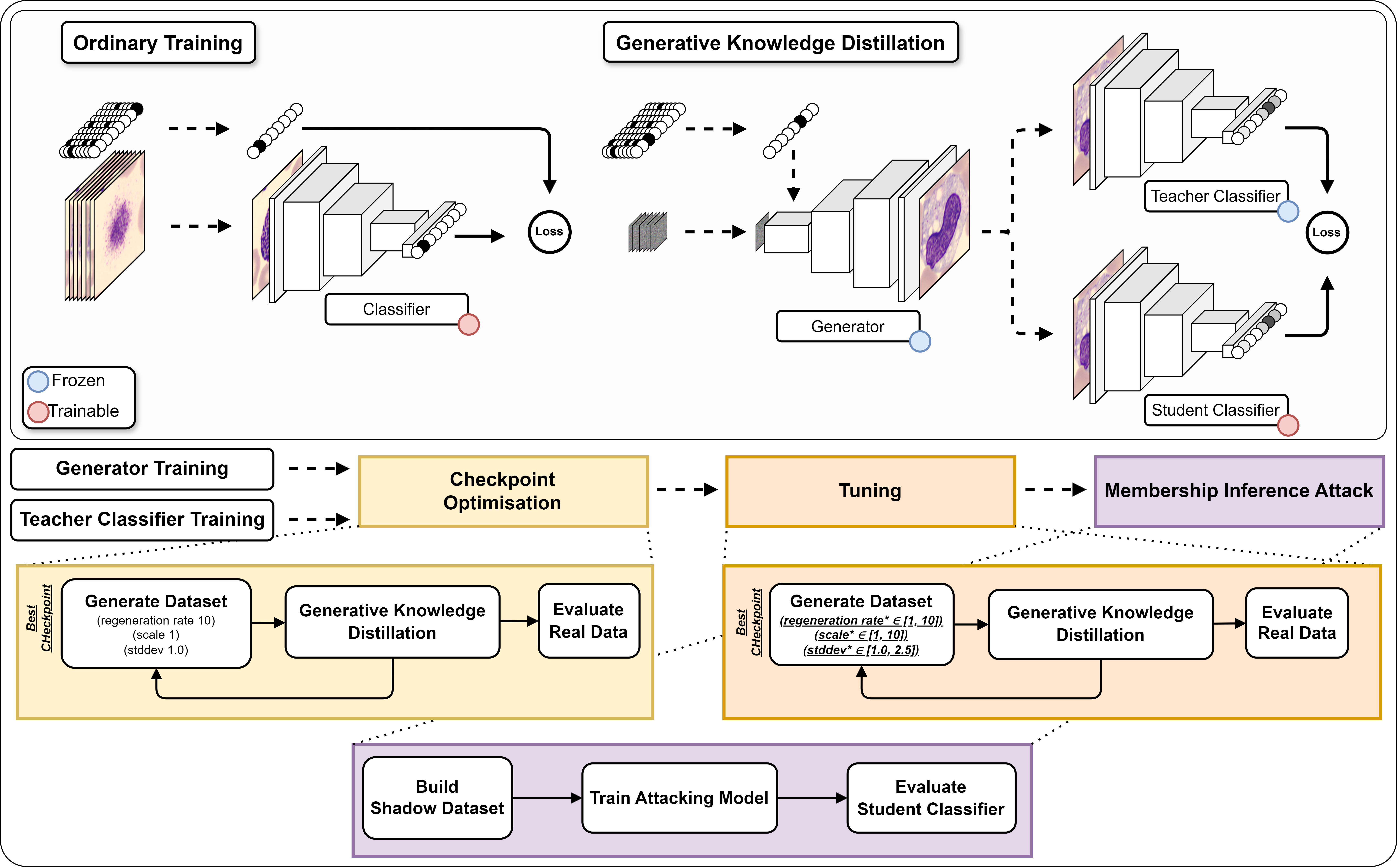}
    \caption{\scriptsize{The difference between an Ordinary Training and the proposed Generative Knowledge Distillation technique, and the illustration of the Knowledge Recycling pipeline.}}
    \label{fig:Pipeline}
\end{figure*}

\noindent This section presents the Knowledge Recycling (KR) pipeline for the creation of synthetic datasets and their subsequent use for training downstream classifiers.
The pipeline starts with a preliminary step where an auxiliary classifier, called \textbf{Teacher Classifier}, and a data generator, called \textbf{Generator}, are trained on the same real dataset.
The first proper step, called \textbf{Checkpoint Optimisation}, aims at identifying the best checkpoint of the Generator.
During this step, a classifier is trained for each checkpoint of the Generator.
These classifiers, called \textbf{Student Classifiers}, have the same architecture as the Teacher Classifier and are trained using the same training technique.
For each checkpoint, the generation of the synthetic datasets is performed using the proposed technique called Generative Knowledge Distillation (GKD), which is explained in detail in the Subsection~\ref{subsec:Checkpoint}.
Once the optimal checkpoint is identified, the \textbf{Tuning} step follows, in which the generation parameters are optimised and the final Student Classifier is trained.
Finally, the last step, called \textbf{Membership Inference Attack}, tests the robustness of the Student Classifier against the homonymous privacy attack. A graphical representation is shown in Figure\ref{fig:Pipeline}.

\subsection{Teacher Classifier}\label{subsec:Teacher}
\noindent The Teacher Classifier plays a key role in the KR pipeline, as it is not only the core of the GKD technique, but also the benchmark against which the Student Classifiers can be compared in terms of accuracy performance and resistance to privacy attacks. 
In order to have a fair and robust comparison, the architecture and training technique of the Teacher Classifier is also replicated for the Student Classifiers. 
Having to balance performance and training speed, since each checkpoint requires a whole one, the chosen architecture is a ResNet14 model~\cite{he2016deep}. Training is done in Mixed Precision for 500 epochs with SGD optimiser, initial learning rate of 0.5, cosine annealing scheduler and TrivialAugment and MixUp as main augmentation~\cite{loshchilov2022sgdr,muller2021trivialaugment,zhang2018mixup}. For more details see \ref{app:Classifier}.

\subsection{Generator}\label{subsec:Image}
\noindent In the field of image generation, Generative Adversarial Networks (GAN) and Denoising Diffusion Probabilistic Models (DDPM) currently represent the state of the art~\cite{goodfellow2014generative,ho2020denoising}. Although they differ significantly in their operation, both approaches offer high and comparable performance in generating different types of media content that can be conditioned in different ways.
GANs are known for their fast inference, but suffer from instability during training. DDPMs, on the other hand, offer more stable training but require longer generation times. Despite recent developments, DDPM models can drastically reduce the number of denoising steps, but their generation times are still too long to compete with GANs in generating large amounts of data~\cite{sauer2024fast}.

\noindent For this work, a GAN-based approach was chosen, which favours the speed of inference. In particular, a modified version of BigGAN-Deep was chosen, a model that represents a milestone in the development of GAN~\cite{brock2018large} models. Indeed, BigGAN introduced several important innovations, including the use of conditional batch normalisation, the use of a truncation trick to control the trade-off between quality and generation diversity, combined with advanced optimisation techniques to handle large networks, such as spectral normalisation~\cite{miyato2018spectral}.
The proposed implementation modifies the original BigGAN-Deep model in several aspects. The hinge loss is replaced by a logistic loss, and the tanh activation is replaced by a sigmoid. In addition, regularisation techniques such as label smoothing have been introduced to improve the quality of the discriminator, and the AdamW optimiser with a weight decay of 0.0005 has been adopted. These changes aim at improving the stability of the training and the quality of the generated images.
The model was trained for 500 epochs, with a 4:1 ratio between discriminator and generator updates. To ensure the robustness of the model, we implemented a system of saving checkpoints at regular intervals of 5 epochs. A detailed description of the implementation and a comparison between the vanilla model and our modified version can be found in \ref{app:Generator}.

\subsection{Evaluation Metric}\label{subsec:Evaluation}
\noindent In the evaluation of image generators, the most widely used metrics in the literature are the Inception Score (IS) and the Fréchet Inception Distance (FID)~\cite{salimans2016improved,heusel2017gans}. The IS aims to quantify the quality of the distribution generated by assessing both the clarity and diversity of the images produced. The FID, on the other hand, provides a more comprehensive measure by comparing the generated distribution with the actual distribution used to train the Generator, thus capturing both the quality and fidelity of the synthetic images.

\noindent However, recent studies have highlighted the limitations of using these metrics to assess the usefulness of generated images in downstream learning contexts. A lack of correlation was observed between IS and FID and the effectiveness of the generated data for subsequent classification tasks~\cite{lampis2023bridging}. Furthermore, a trade-off between the quality of individual images and the diversity of the generated distribution was identified~\cite{astolfi2024consistency}. 
The maximisation of IS and FID tends to favour the quality of the generated images at the expense of the variety, which is crucial for the creation of synthetic datasets that favour the generalisation of the models trained on them.

\noindent In this study, the Classification Accuracy Score (CAS) is adopted as the main metric. The CAS measures the validation accuracy on real data of a classifier trained on synthetic datasets~\cite{ravuri2019classification}. 
This metric helps to identify the training epoch that produces the most effective synthetic dataset and, like IS and FID, helps to prevent the mode collapse of the Generator.

\subsection{Checkpoint Optimisation}\label{subsec:Checkpoint}
\noindent Once both the Teacher Classifier and the Generator have been defined, trained on real training data and frozen, it is possible to proceed to the Checkpoint Optimisation step.
The goal of this first step is to identify the optimal checkpoint to maximise the performance of the downstream Student Classifier models.
For each Generator checkpoint, a Student Classifier is trained using a strategy similar to that of the Teacher Classifier, but with a reduced number of epochs -- 100 instead of 500 -- for efficiency reasons. At the beginning of each training session, a synthetic dataset of the same cardinality as the real one is generated using the current checkpoint. The input noise is sampled from a multivariate Gaussian distribution with a standard deviation of 1.0. To maintain data diversity, the synthetic data set is fully regenerated every 10 epochs during training.

\noindent Previous studies have demonstrated the effectiveness of filtering the generated data to improve CAS. Dat et al. used a model similar to the Teacher Classifier to exclude images with inconsistent predictions~\cite{dat2019classifier}, while Lampis et al. introduced an additional filtering step based on the confidence of the predictions~\cite{lampis2023bridging}.

\noindent In the KR pipeline, the technique of \textbf{Generative Knowledge Distillation} (GKD) is proposed and adopted. In contrast to filtering methods, the Teacher Classifier is used to evaluate the generated images and produce soft labels for the Student Classifier. These probability labels are more informative than binary ones, as they capture uncertainties and correlations between classes, leading to a significant improvement in CAS, as detailed in \ref{app:Training}.
This approach optimises both the quality of the information passed to the Student Classifier and the efficiency of the synthetic dataset generation process, allowing the desired dataset cardinality to be achieved faster than with filter-based techniques.

\subsection{Tuning}\label{subsec:Tuning}
\noindent After the identification of the optimal checkpoint with respect to the CAS metric, it is possible to proceed with the Tuning step to optimise also the generation parameters. These parameters, held constant during the Checkpoint Optimisation step, are now re-computed to further improve Student Classifiers performance.
The parameters being optimised are:
\begin{itemize}
    \item The regeneration rate of the synthetic dataset: previously fixed at 10 epochs, is now varied between 1 and 10 epochs.
    \item The scale of the cardinality of the generated dataset: previously set at 1, is now made to vary between 1 and 10.
    \item The standard deviation used during sampling from the multivariate Gaussian distribution: previously equal to 1.0, is now made to vary between 1.0 and 2.5.
\end{itemize}
\noindent It has been shown in previous works that regenerating the dataset more frequently and creating more numerous datasets contributes to the improvement of CAS~\cite{lampis2023bridging,lomurno2024stable}. With regard to standard deviation, this approach, in the opposite direction to the Truncation Trick implemented in the BigGAN-Deep vanilla model, aims to favour a more varied generation, even at the expense of the perceptual quality of the generated data~\cite{brock2018large,lampis2023bridging}.
The Tuning step is carried on via a Tree-structured Parzen Estimator associated with a Hyperband pruning mechanism~\cite{bergstra2011algorithms,li2018hyperband}. The optimisation proceeds for 50 iterations. In each iteration, a Student Classifier is trained with the same procedure used in Checkpoint Optimisation but using the data generated with the current parameter configuration, with the aim of maximising CAS.
At the end of the search, the optimal parameter configuration is used to train the final Student Classifier for 500 epochs.

\subsection{Membership Inference Attack}\label{subsec:Mia}
\noindent The final step in the KR pipeline involves a resistance test for the Student Classifier against a Membership Inference Attack (MIA). This type of attack aims to compromise privacy by identifying the training data stored within the attacked model. In the context of this study, sensitive training data is never directly exposed to the Student Classifier, but is only used for training the Generator and Teacher Classifier.
The objective of this step is therefore to assess the effectiveness of the MIA in identifying the data used to train the Generator via the Student Classifier and to compare this effectiveness with that of the same attack carried out against the Teacher Classifier.
To perform the MIA, the shadow models technique proposed by Shokri et al. is adopted~\cite{shokri2017membership}..
The implementation involves:
\begin{enumerate}
    \item Creation of 10 shadow models, identical in architecture and training technique to the Teacher Classifier.
    \item Utilisation of the validation dataset to train each shadow model, split with 45/10/45 splits to simulate the training, validation and test sets, mixed before splitting to obtain different splits for each shadow model.
    \item Shadow dataset generation: training and test data are given as input to each shadow model. The logits obtained represent the shadow dataset features, while a binary label indicates whether they belong to the training set or the external (test) set.
    \item Division of the dataset according to the classes of the original dataset with which the shadow models were trained.
    \item Training of three models (Logistic Regression, Support Vector Classifier with RBF, Random Forest) for each class of the dataset.
    \item Selection of the best performing model for each class, based on the Area Under the Receiver Operating Characteristic Curve (AUC) metric.
\end{enumerate}
The overall attack model is composed of the best classifier for each class in the dataset and is used against both the Teacher Classifier and the Student Classifier.
The evaluation of resistance to MIAs is based on two metrics: the AUC, typically used to evaluate this type of attack, and the Accuracy Over Privacy (AOP), which provides an estimate of the trade-off between performance -- measured as test accuracy -- and resistance to MIAs~\cite{lomurno2023discriminative}.
The objective of this step is to examine whether the proposed training from synthetic data may constitute an additional layer of privacy, making the attack less effective.

\begin{figure*}[p]
    \centering
    \includegraphics[width=1.\linewidth]{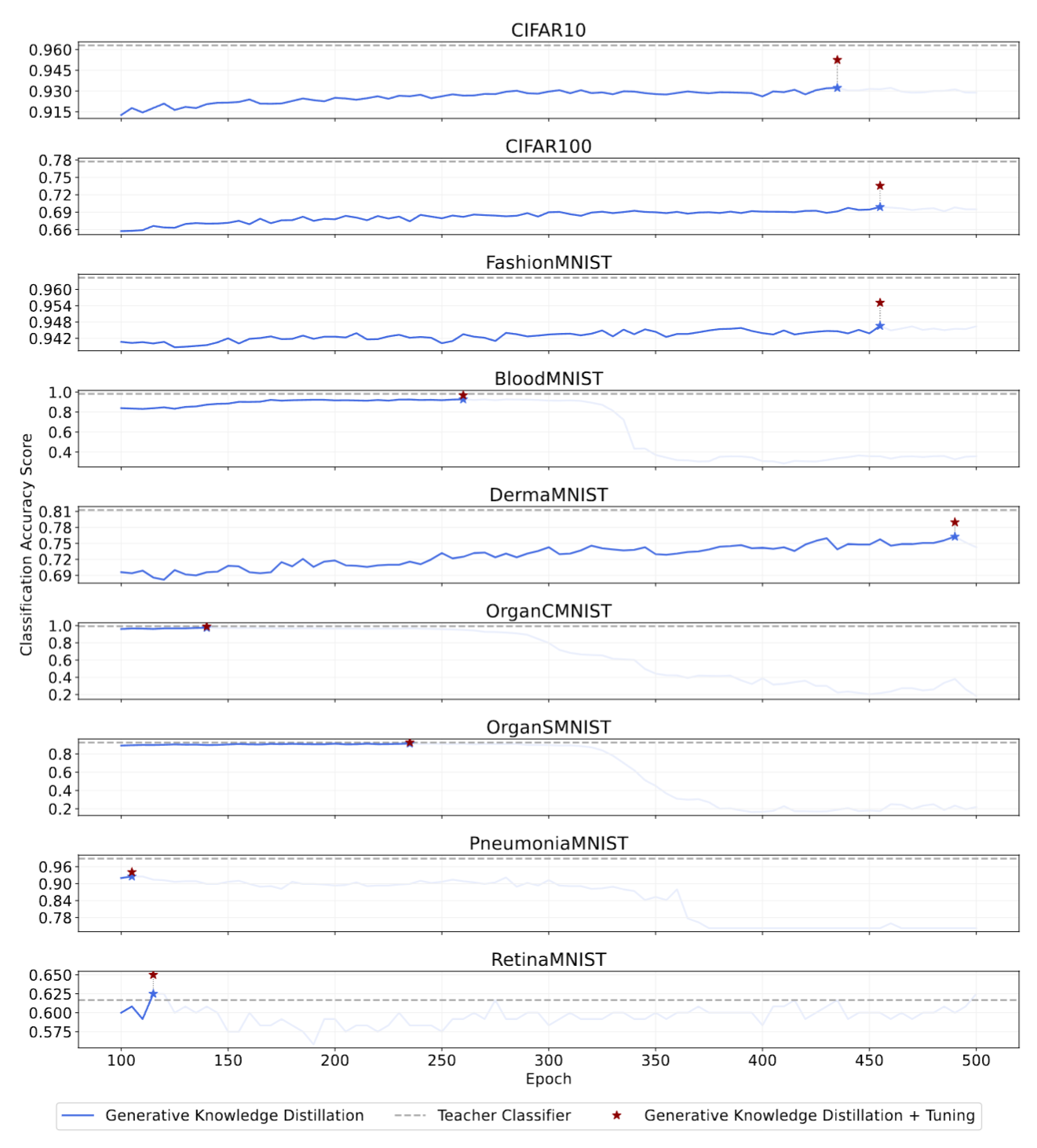}
    \caption{\scriptsize{The Classification Accuracy Score (CAS) of the validation calculated for each checkpoint of the Generator for the considered datasets. The continuous blue line represents the CAS obtained during the Checkpoint Optimisation step using the Generative Knowledge Distillation technique. The best checkpoint is marked with a blue star. The dashed grey line represents the best validation Accuracy obtained with the Teacher Classifier. The red star indicates the optimal checkpoint CAS of the validation after training with Generative Knowledge Distillation with parameters found during the Tuning step.}}
    \label{fig:Checkpoint}
\end{figure*}

\section{Experiments and Results}\label{sec:Results}
\noindent The experiments were performed on nine image datasets, all rescaled to 32x32. CIFAR10, CIFAR100 and FashionMNIST were used both for the final comparisons and to calibrate and test the Knowledge Recycling (KR) pipeline, as described in detail in \ref{app:Generator} and \ref{app:Training}. The six medical datasets - BloodMNIST, DermaMNIST, OrganCMNIST, OrganSMNIST, PneumoniaMNIST and RetinaMNIST - contain real images from the MedMNIST v2 benchmark~\cite{medmnistv2}. These medical datasets represent the primary field of application for the proposed technique. The KR pipeline, having been calibrated on the three aforementioned datasets, is subsequently applied to these medical datasets without further specific adaptations. This approach allows for the evaluation of the technique's effectiveness and robustness in a more specialised and complex context, distinct from that on which it was initially calibrated. Further details of all the datasets used can be found at \ref{app:Datasets}. Experiments were run on 4 NVIDIA Quadro RTX 6000 GPUs.

\begin{table*}[t]
    \centering
    \setlength{\tabcolsep}{3pt}
    \caption{\scriptsize{The optimum generation parameters identified after the Tuning step for each dataset under consideration. The $\Delta$ CAS column represents the improvement of the validation Classification Accuracy Score compared to the performance obtained from the default generation parameters.}}
    \vspace{4pt}
    \footnotesize
    \begin{tabularx}{\textwidth}{l*{4}{Y}}
        \toprule
        \scriptsize{Hyperparameter} & \scriptsize{Standard Deviation} & \scriptsize{Regeneration Rate} & \scriptsize{Cardinality Scale} & \scriptsize{$\Delta$ CAS}\\ 
        \midrule
        \scriptsize{CIFAR10}         &  1.40 &  9 &  8 & +2.02\\
        \scriptsize{CIFAR100}        &  1.44 &  7 &  9 & +3.67\\
        \scriptsize{FashionMNIST}    &  1.58 &  1 &  6 & +0.85\\
        \midrule[0.05pt]
        \scriptsize{BloodMNIST}      &  2.23 &  1 &  10 & +4.03\\
        \scriptsize{DermaMNIST}      &  1.23 &  10 &  8 & +2.69\\
        \scriptsize{OrganCMNIST}     &  2.33 &  2 &  10 & +1.42\\
        \scriptsize{OrganSMNIST}     &  2.42 &  7 &  10 & +1.22\\
        \scriptsize{PneumoniaMNIST}  &  2.15 &  3 &  5 & +1.53\\
        \scriptsize{RetinaMNIST}     &  1.61 &  2 &  7 & +2.50\\
        \bottomrule
    \end{tabularx}
    \label{tab:Tuning}
\end{table*}

\noindent The Figure~\ref{fig:Checkpoint} illustrates the results of the Checkpoint Optimisation and Tuning phases, expressed as Classification Accuracy Score (CAS) on the respective validation sets, compared with the optimal Accuracy performance of the Teacher Classifier on the same set.
The importance of the selection of the optimal checkpoint and its evaluation via CAS is evident both for Generators with more stable checkpoints (e.g. DermaMNIST, RetinaMNIST) and for those subject to mode collapse and consequent drop in performance (e.g. BloodMNIST, OrganSMNIST).
The application of the Generative Knowledge Distillation (GKD) technique alone demonstrates to be sufficient to obtain results close to those of the Teacher Classifier. In the case of the RetinaMNIST dataset, a more accurate model is even obtained from the synthetic data alone.
The Tuning step turns out to be beneficial overall, increasing the validation CAS from a minimum of 0.85\% for FashionMNIST to a maximum of 4.03\% for BloodMNIST, as reported in Table~\ref{tab:Tuning}.
This improvement is due to two factors. The first is the increased availability of information due to the higher cardinality of the generated datasets and their higher recycling frequency. The second is the increased diversity of data due to sampling with a larger standard deviation, which, in combination with the GKD technique, also makes it possible to exploit images that would be uninformative if associated with the hard label used to generate them. These images would likely be filtered out and discarded if coupled with another training technique from synthetic data.

\noindent The Table~\ref{tab:Final} presents the results of the final comparison between Teacher Classifier and Student Classifier. The testing CAS of the Student Classifiers approaches the testing Accuracy of the Teacher Classifiers on average, exceeding it in the cases of PneumoniaMNIST and RetinaMNIST.
With regard to resilience to Membership Inference Attacks (MIA), the Student Classifiers demonstrate greater resilience, with performance for the attacker close to random guessing.
The Accuracy Over Privacy (AOP) metric, which measures the trade-off between performance and resilience to MIAs, shows that Student Classifiers consistently outperform Teacher Classifiers. This implies that the slight margin of loss in CAS on thest set is positively compensated by the increased and almost total resilience to MIAs.

\begin{table*}[t]
    \centering
    \setlength{\tabcolsep}{3pt}
    \caption{\scriptsize{The comparison of Accuracy, AUC$_{MIA}$ and AOP performance between Teacher Classifier and Student Classifier calculated on the test set -- for the Student Classifier the Accuracy is intended as Classification Accuracy Score. The best score is highlighted with \textbf{bold}.}}
    \vspace{4pt}
    \footnotesize
    \begin{tabularx}{\textwidth}{l*{6}{Y}}
        \toprule
        & \multicolumn{2}{c}{\scriptsize{Accuracy $\uparrow$}} & \multicolumn{2}{c}{\scriptsize{AUC$_{MIA}$ $\downarrow$}} & \multicolumn{2}{c}{\scriptsize{AOP $\uparrow$}}\\ 
        \cmidrule(lr){2-3} \cmidrule(lr){4-5} \cmidrule(lr){6-7}
        \scriptsize{Model} & \scriptsize{Teacher Classifier} & \scriptsize{Student Classifier} & \scriptsize{Teacher Classifier} & \scriptsize{Student Classifier} & \scriptsize{Teacher Classifier} & \scriptsize{Student Classifier} \\ 
        \midrule
        \scriptsize{CIFAR10}         &  \textbf{96.24} &  95.83 &  56.22 &  \textbf{51.21} &  76.13 &  \textbf{91.34} \\
        \scriptsize{CIFAR100}        &  \textbf{77.65} &  74.92 &  63.75 &  \textbf{52.31} &  47.77 &  \textbf{68.44} \\
        \scriptsize{FashionMNIST}    &  \textbf{95.94} &  95.21 &  52.69 &  \textbf{50.50} &  86.38 &  \textbf{93.34} \\
        \midrule[0.05pt]
        \scriptsize{BloodMNIST}      &  \textbf{97.49} &  96.26 &  51.20 &  \textbf{50.38} &  92.96 &  \textbf{94.81} \\
        \scriptsize{DermaMNIST}      &  \textbf{79.50} &  76.46 &  59.41 &  \textbf{50.43} &  56.31 &  \textbf{75.15} \\
        \scriptsize{OrganCMNIST}     &  \textbf{93.16} &  90.23 &  55.73 &  \textbf{53.07} &  74.98 &  \textbf{80.11} \\
        \scriptsize{OrganSMNIST}     &  \textbf{79.78} &  78.76 &  54.29 &  \textbf{52.38} &  67.66 &  \textbf{71.78} \\
        \scriptsize{PneumoniaMNIST}  &  86.54 &  \textbf{86.70} &  50.39 &  \textbf{50.00} &  85.19 &  \textbf{86.70} \\
        \scriptsize{RetinaMNIST}     &  54.25 &  \textbf{55.00} &  52.36 &  \textbf{50.58} &  49.48 &  \textbf{53.74} \\
        \midrule
        \scriptsize{Min Imp}         &  - &  -3.04 &  - &  -0.39  &  - &  1.51\\
        \scriptsize{Avg Imp}         &  - &  -1.24 &  - &  -3.90  &  - &  7.72\\
        \scriptsize{Max Imp}         &  - &   0.75 &  - &  -11.40 &  - &  20.66\\
        \bottomrule
    \end{tabularx}
    \label{tab:Final}
\end{table*}

\section{Discussion and Limitations}\label{sec:Discussion}
\noindent The Knowledge Recycling (KR) technique proposed in this study has been shown to be effective in creating Student Classifiers with comparable performance to the corresponding Teacher Classifiers, while maintaining considerable resistance to Membership Inference Attacks (MIA). This approach, initially calibrated on standard datasets such as CIFAR10, CIFAR100 and FashionMNIST, and subsequently applied to six medical image datasets from the MedMNIST v2 benchmark, establishes a new state-of-the-art in this field.

\noindent The average performance gap between Teacher Classifiers and Student Classifiers was reduced to -1.24\% in terms of the Classification Accuracy Score (CAS) over the test sets, a significant improvement on previous results. This progress is particularly remarkable considering the use of a single Generator, in contrast to previous works. Dat et al. had achieved an average gap of -10.08\% with a single Generator and -5.81\% with six, while Lampis et al. had achieved -3.87\% with a single Generator and -2.63\% with six~\cite{lampis2023bridging}. The approach proposed in this study exceeds these results, suggesting potential for improvement through the use of multiple Generators in parallel.

\noindent The inclusion of a metric to empirically measure one of the privacy-related aspects, such as resistance to MIAs, proved to be crucial for a richer and more multifaceted evaluation of the proposed method, especially if the data under analysis are medical images with potential sensitivities to violations of their privacy. Teacher Classifiers, trained with regularization and augmentation techniques, showed partial resistance to MIAs, confirming the privacy properties associated with such techniques~\cite{lomurno2022utility}. However, Student Classifiers showed almost complete resistance to these attacks, showing a significant privacy advantage of the proposed approach.

\noindent The main limitations of this study concern the small size of the images used (32x32 pixels) and the choice of models that are efficient but not comparable in performance with the current state of the art in their respective tasks. 
These decisions were dictated by computational efficiency considerations, given the onerous nature of the KR pipeline. 
The use of higher resolution images and more complex models, both for the Classifier (ResNet14) and the Generator (BigGAN-Deep), could lead to further performance improvements. In particular, upgrading the Generator model could further reduce the performance gap between Teacher and Student Classifiers, potentially outperforming the Teacher.
The scalability of the proposed approach, both in terms of the number of Generators and the cardinality and frequency of generation, offers exciting prospects for future developments. With continued hardware advancement, it is plausible that in the near future it will be possible to apply this technique with more complex models and on larger datasets, opening up new possibilities in the field of private learning and the generation of high-quality synthetic data.

\section{Conclusions}\label{sec:Conclusion}
\noindent In this paper, the Knowledge Recycling pipeline was presented, demonstrating how synthetic data can be generated and used to train downstream classifiers.
It has been shown how the Generative Knowledge Distillation technique, used within the pipeline, improves the quality of information transferable to such downstream classifiers compared to techniques previously proposed in the literature.
It was possible to simultaneously reduce the gap between the performance obtainable from real data alone and that obtainable from generated data, setting a new state of the art, as well as to obtain models from synthetic data that manifest privacy properties such that Membership Inference Attacks are ineffective.
This was tested on real medical image datasets, demonstrating how it is possible to simultaneously preserve performance and reduce privacy attack surfaces.

\section{Acknowledgements}\label{sec:Acknowledgements}
\noindent This paper is supported by the FAIR (Future Artificial Intelligence Research) project, funded by the NextGenerationEU program within the PNRR-PE-AI scheme (M4C2, investment 1.3, line on Artificial Intelligence).

\bibliographystyle{unsrt}  
\bibliography{references}

\newpage
\section*{Appendix A - Image Generator}\label{app:Generator}
\begin{figure*}[t]
    \centering
    \includegraphics[width=1\linewidth]{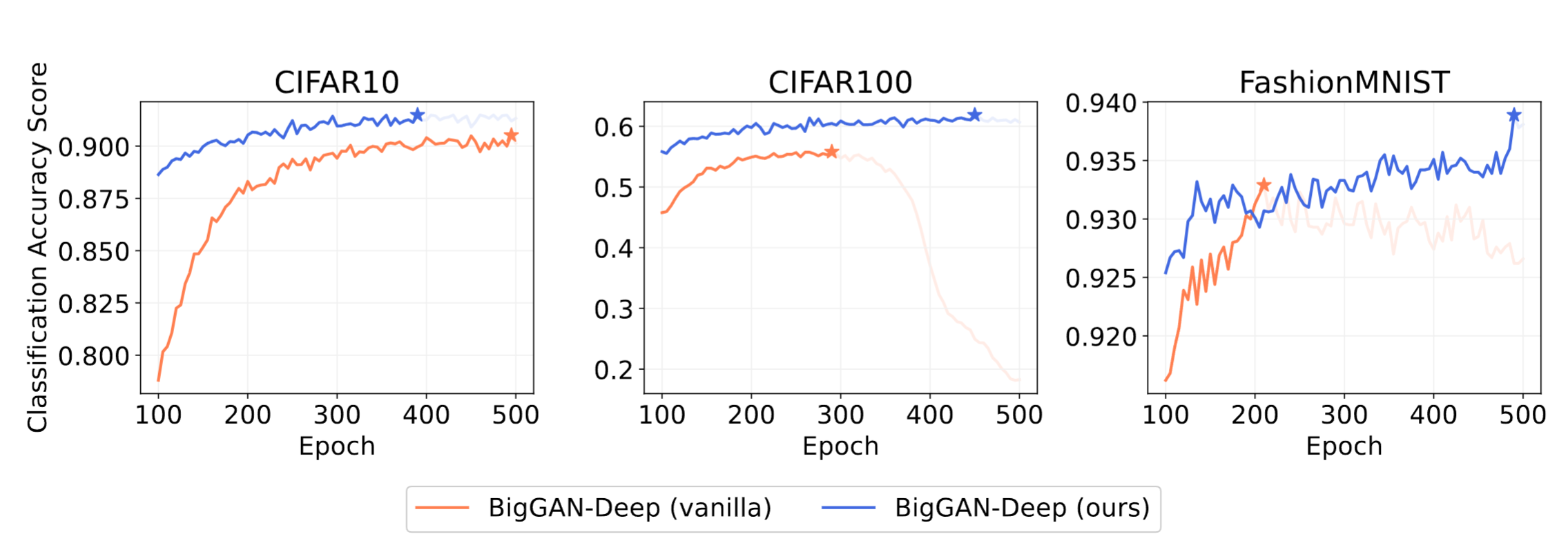}
    \footnotesize
    \caption{\footnotesize{The Classification Accuracy Score (CAS) of the validation calculated for each checkpoint of the BigGAN-Deep (vanilla) and BigGAN-Deep (ours) generators for the considered datasets.}}
    \label{fig:Generators}
\end{figure*}

\noindent This section shows and compares the BigGAN-Deep (vanilla) model and its modified version used in this study, called BigGAN-Deep (ours). 
Table~\ref{tab:Generators} shows the main changes between the two versions.
Figure~\ref{fig:Generators} shows the validation Classification Accuracy Score calculated for each checkpoint for the CIFAR10, CIFAR100 and FashionMNIST datasets. Each model is trained in the same way as during the Chackpoint Optimisation step of the Knowledge Recycling pipeline, with the exception of the generation of the synthetic dataset, which is not regenerated during training and whose labels correspond to the hard labels used to condition the generation.

\begin{table*}[!]
    \centering
    \setlength{\tabcolsep}{3pt}
    \caption{\footnotesize{The comparison between the parameters of the BigGAN-Deep (vanilla) model and its modified version used in this work and called BigGAN-Deep (ours).}}
    \vspace{4pt}
    \footnotesize
    \begin{tabularx}{\textwidth}{lYY}
        \toprule
          \footnotesize{Model} & \footnotesize{BigGan-Deep (vanilla)} &  \footnotesize{BigGan-Deep (ours)}\\ 
          \midrule

          \footnotesize{Latent dimension} & \footnotesize{128} &  \footnotesize{128}\\
          \footnotesize{Shared dimension} & \footnotesize{128} &  \footnotesize{128}\\
          \footnotesize{Batch size} & \footnotesize{64} &  \footnotesize{64}\\ 
          \footnotesize{Adversarial loss} & \footnotesize{Hinge} &  \footnotesize{Logistic}\\ 
          \footnotesize{Precision} & \footnotesize{Full} & \footnotesize{Mixed}\\
           &  \\
          \footnotesize{G conditioning} & \footnotesize{cBN} &  \footnotesize{cBN}\\ 
          \footnotesize{D conditioning} & \footnotesize{PD} &  \footnotesize{PD}\\ 
          
          \footnotesize{G spectral normalisation} & \footnotesize{True} &  \footnotesize{True}\\ 
          \footnotesize{D spectral normalisation} & \footnotesize{True} &  \footnotesize{True}\\ 
          
          \footnotesize{G optimiser} & \footnotesize{Adam} &  \footnotesize{Adam}\\ 
          \footnotesize{D optimiser} & \footnotesize{Adam} &  \footnotesize{AdamW}\\ 

          \footnotesize{G learning rate} & \footnotesize{0.0002} &  \footnotesize{0.0002}\\ 
          \footnotesize{D learning rate} & \footnotesize{0.0002} &  \footnotesize{0.0002}\\ 

          \footnotesize{G beta1} & \footnotesize{0.5} &  \footnotesize{0.5}\\ 
          \footnotesize{G beta2} & \footnotesize{0.999} &  \footnotesize{0.999}\\ 
          \footnotesize{D beta1} & \footnotesize{0.5} &  \footnotesize{0.5}\\ 
          \footnotesize{D beta2} & \footnotesize{0.999} &  \footnotesize{0.999}\\ 
          
          \footnotesize{G ema} & \footnotesize{True} &  \footnotesize{True}\\ 
          \footnotesize{G ema decay} & \footnotesize{0.9999} &  \footnotesize{0.9999}\\ 
          \footnotesize{G ema starting step} & \footnotesize{1000} &  \footnotesize{1000}\\ 

          \footnotesize{D basket size} & \footnotesize{3} &  \footnotesize{1}\\ 
          \footnotesize{D label smoothing} & \footnotesize{-} &  \footnotesize{0.1}\\ 
          \footnotesize{D weight decay} & \footnotesize{-} &  \footnotesize{0.004}\\ 
          \footnotesize{D updates per step} & \footnotesize{2} &  \footnotesize{4}\\ 
           &  \\
          \footnotesize{G attention resolution} & \footnotesize{16} &  \footnotesize{16}\\ 
          \footnotesize{D attention resolution} & \footnotesize{16} &  \footnotesize{16}\\ 

          \footnotesize{G convolutional dimension} & \footnotesize{128} &  \footnotesize{128}\\ 
          \footnotesize{D convolutional dimension} & \footnotesize{128} &  \footnotesize{128}\\ 
          
          \footnotesize{G depth} & \footnotesize{2} &  \footnotesize{2}\\ 
          \footnotesize{D depth} & \footnotesize{2} &  \footnotesize{2}\\ 
          
          \footnotesize{G model size} & \footnotesize{41M} &  \footnotesize{41M}\\ 
          \footnotesize{D model size} & \footnotesize{39M} &  \footnotesize{39M}\\ 

        \bottomrule
    \end{tabularx}
    \label{tab:Generators}
\end{table*}

\newpage
$ $
\newpage
\section*{Appendix B - Image Classifier}\label{app:Classifier}
\noindent This section shows the implementation details of the Classifiers used in this work.
Within the Table~\ref{tab:Classifier}, the configuration, training and data augmentation parameters are shown in detail.
\begin{table*}[t]
    \centering
    \setlength{\tabcolsep}{3pt}
    \caption{\footnotesize{The main configuration, training and data augmentation parameters used to realise each Classifier in this work.}}
    \vspace{4pt}
    \footnotesize
    \begin{tabularx}{\textwidth}{lY}
        \toprule
        \footnotesize{Model} & \footnotesize{ResNet14}\\ 
        \midrule
        \footnotesize{Initial filters} & \footnotesize{64}\\
        \footnotesize{Batch size} & \footnotesize{256}\\
        \footnotesize{Loss} & \footnotesize{Categorical cross-entropy}\\
        \footnotesize{Precision} & \footnotesize{Mixed}\\
        
        \footnotesize{Learning rate} & \footnotesize{0.5}\\
        \footnotesize{Learning rate scheduler} & \footnotesize{Cosine annealing}\\
        \footnotesize{Warm-up epochs} & \footnotesize{10}\\
        \footnotesize{Warm-up learning rate} & \footnotesize{0.00001}\\
        
        &  \\
        \footnotesize{Optimiser} & \footnotesize{SGD}\\
        \footnotesize{Momentum} & \footnotesize{0.9}\\
        \footnotesize{Weight decay} & \footnotesize{0.0005}\\
        \footnotesize{Nesterov} & \footnotesize{True}\\
        \footnotesize{Clipnorm} & \footnotesize{1.0}\\
        \footnotesize{Label smoothing} & \footnotesize{0.1}\\
        &  \\
        \footnotesize{TrivialAugment} & \footnotesize{True}\\
        \footnotesize{TrivialAugment interpolation} & \footnotesize{Bilinear}\\
        \footnotesize{MixUp} & \footnotesize{True}\\
        \footnotesize{MixUp alpha} & \footnotesize{0.2}\\
        \footnotesize{Random horizontal flip} & \footnotesize{True}\\
        \footnotesize{Padding} & \footnotesize{2}\\
        \footnotesize{Random crop} & \footnotesize{True}\\
        \bottomrule
    \end{tabularx}
    \label{tab:Classifier}
\end{table*}

\newpage
\section*{Appendix C - Datasets}\label{app:Datasets}
\noindent This section shows the details of the datasets used in this work.
Table~\ref{tab:Datasets} shows how each dataset has been acquired, the number of classes, the cardinality of the samples and the size of each of the training, validation and test splits.
\begin{table*}[!]
    \centering
    \setlength{\tabcolsep}{3pt}
    \caption{\footnotesize{Information on the datasets used in this work.}}
    \vspace{4pt}
    \footnotesize
    \begin{tabularx}{\textwidth}{lcYYc}
        \toprule
        \footnotesize{Dataset} & \footnotesize{Data Modality} & \footnotesize{\# Classes} & \footnotesize{\# Samples} & \footnotesize{\# Training/Validation/Test}\\ 
        \midrule
        \footnotesize{CIFAR10} & \footnotesize{Camera - Various Subjects} & \footnotesize{10} & \footnotesize{60000} & \footnotesize{40000/10000/10000}\\
        \footnotesize{CIFAR100} & \footnotesize{Camera - Various Subjects} & \footnotesize{100} & \footnotesize{60000} & \footnotesize{40000/10000/10000}\\
        \footnotesize{FashionMNIST} & \footnotesize{Camera - Clothing} & \footnotesize{10} & \footnotesize{70000} & \footnotesize{50000/10000/10000}\\
        \midrule
        \footnotesize{BloodMNIST} & \footnotesize{Blood Cell Microscope} & \footnotesize{8} & \footnotesize{17092} & \footnotesize{11959/1712/3421}\\
        \footnotesize{DermaMNIST} & \footnotesize{Dermatoscope} & \footnotesize{7} & \footnotesize{10015} & \footnotesize{7007/1003/2005}\\
        \footnotesize{OrganCMNIST} & \footnotesize{Abdominal CT} & \footnotesize{11} & \footnotesize{23583} & \footnotesize{12975/2392/8216}\\
        \footnotesize{OrganSMNIST} & \footnotesize{Abdominal CT} & \footnotesize{7} & \footnotesize{10015} & \footnotesize{13932/2452/8827}\\
        \footnotesize{PneumoniaMNIST} & \footnotesize{Chest X-Ray} & \footnotesize{7} & \footnotesize{5856} & \footnotesize{4708/524/624}\\
        \footnotesize{RetinaMNIST} & \footnotesize{Fundus Camera} & \footnotesize{7} & \footnotesize{1600} & \footnotesize{1080/120/400}\\
        \bottomrule
    \end{tabularx}
    \label{tab:Datasets}
\end{table*}

\newpage
\section*{Appendix D - Training Strategies}\label{app:Training}
\noindent This section presents a comparative analysis of the performance of the different training strategies examined in the preliminary phase of this study.
The Figure~\ref{fig:Generators} illustrates the validation Classification Accuracy Score calculated for each checkpoint of the BigGAN-Deep (ours) model for the CIFAR10, CIFAR100 and FashionMNIST datasets. The training of each model follows a similar procedure to that used during the Checkpoint Optimisation step of the Knowledge Recycling pipeline, with the exception of the generation of the synthetic dataset.

\noindent For the Baseline strategy, the dataset is generated once at the beginning of each training, with a cardinality equal to that used to train the Generator. The Gap Filler strategy applies the filtering technique proposed by Lampis et al. The synthetic dataset is regenerated every 10 epochs, maintaining the same size as that used for the Generator~\cite{lampis2023bridging}. Finally, the Generative Knowledge Distillation (GKD) strategy introduced in this work regenerates the synthetic dataset every 10 epochs, maintaining the same size.
The analysis of the results highlights the superiority of the GKD approach, showing how this methodology significantly increases the amount of information in the generated synthetic datasets compared to the alternatives considered.



\begin{figure*}[t]
    \centering
    \includegraphics[width=1\linewidth]{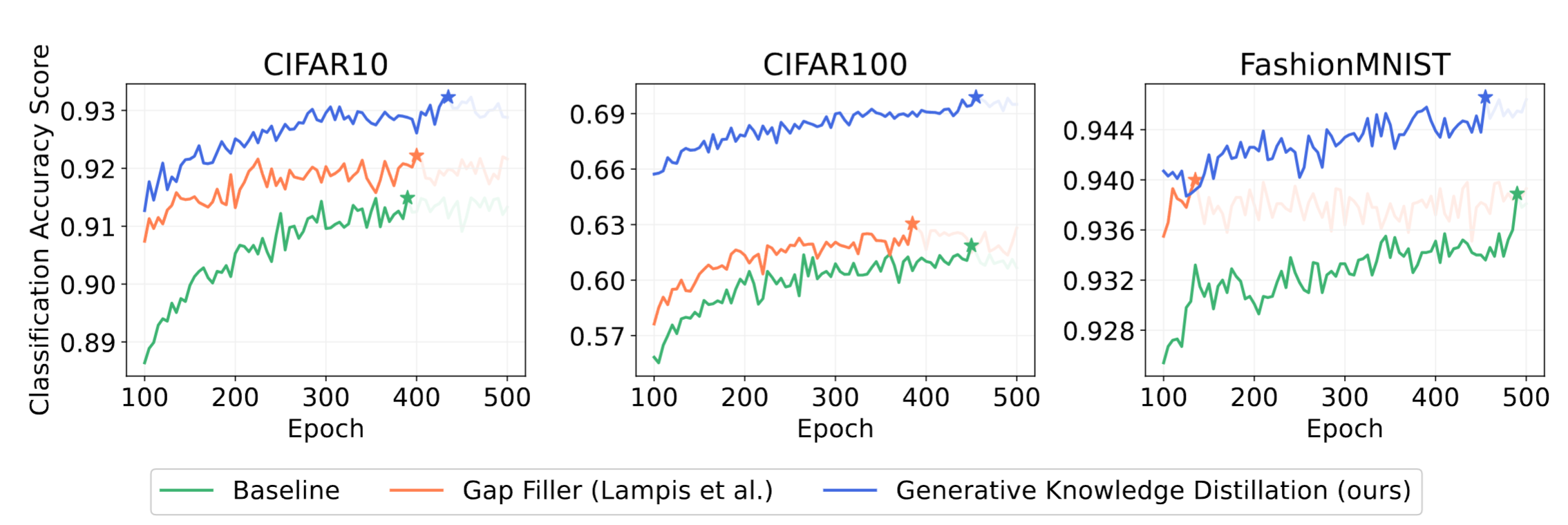}
    \caption{\footnotesize{The Classification Accuracy Score (CAS) of the validation calculated for each checkpoint of the BigGAN-Deep (ours) generator for the considered datasets. The comparison is made between an "ordinary" strategy with a single dataset generation at the beginning of each training (Baseline), the approach presented by Lampis et al. (Gap Filler), and the one proposed in this work (Generative Knowledge Distillation).}}
    \label{fig:Strategy}
\end{figure*}

\end{document}